%% file: main.tex
\documentclass{article}

\PassOptionsToPackage{numbers, compress}{natbib}

\usepackage[final]{neurips_2024}

\usepackage[utf8]{inputenc} 
\usepackage[T1]{fontenc}    
\usepackage{hyperref}       
\usepackage{url}            
\usepackage{booktabs}       
\usepackage{amsfonts}       
\usepackage{nicefrac}       
\usepackage{microtype}      
\usepackage{xcolor}         
\usepackage{graphicx}
\usepackage{subcaption}
\usepackage{tabu}
\usepackage{amsmath}
\usepackage{wrapfig}

\usepackage{amssymb}
\usepackage{pifont}

\newcommand{\topic}[1]
{
\vspace{2mm}\noindent\textbf{#1}
}
\usepackage{caption}
\usepackage{xspace}

\makeatletter
\DeclareRobustCommand\onedot{\futurelet\@let@token\@onedot}
\def\@onedot{\ifx\@let@token.\else.\null\fi\xspace}
\newcommand{\etal}{\emph{et al}\onedot}
\newcommand{\eg}{\emph{e.g}\onedot}
\newcommand{\ie}{\emph{i.e}\onedot}

\newcommand{\cameraready}[1]{\textcolor{black}{#1}}

\title{Rethinking Score Distillation as a \\ Bridge Between Image Distributions }

\author{
{David McAllister$^{1}$\thanks{Equal contribution.} ~ Songwei Ge$^{2*}$ ~ Jia-Bin Huang{$^{2}$} ~ David W. Jacobs{$^{2}$}} \\ \textbf{Alexei A. Efros{$^{1}$ ~ Aleksander Holynski{$^{1}$} ~ Angjoo Kanazawa{$^{1}$}}} \\
\normalsize
$^{1}$\	UC Berkeley  
$^{2}$\ University of Maryland \\
\url{https://sds-bridge.github.io/} \\
}

\begin{document}

\maketitle


\vspace{-15pt}
\begin{abstract}
\vspace{-5pt}
Score distillation sampling (SDS) has proven to be an important tool, enabling the use of large-scale diffusion priors for tasks operating in data-poor domains. 
Unfortunately, SDS has a number of characteristic artifacts that limit its usefulness in general-purpose applications. 
In this paper, we make progress toward understanding the behavior of SDS and its variants by viewing them as solving an optimal-cost transport path from a source distribution to a target distribution. Under this new interpretation, these methods seek to transport corrupted images (source) to the natural image distribution (target).
We argue that current methods' characteristic artifacts are caused by (1) linear approximation of the optimal path and (2) poor estimates of the source distribution.
We show that calibrating the text conditioning of the source distribution can produce high-quality generation and translation results with little extra overhead.
Our method can be easily applied across many domains, matching or beating the performance of specialized methods. 
We demonstrate its utility in text-to-2D, text-based NeRF optimization, translating paintings to real images, optical illusion generation, and 3D sketch-to-real. 
We compare our method to existing approaches for score distillation sampling and show that it can produce high-frequency details with realistic colors.
\end{abstract}

\input{sections/1_introduction}
\input{sections/2_related_work}
\input{sections/3_method}
\input{sections/4_experiments}

\input{sections/4.5_discussion}
\input{sections/5_conclusion}

\paragraph{Acknowledgment.} We thank Matthew Tancik, Jiaming Song, Riley Peterlinz, Ayaan Haque, Ethan Weber, Konpat Preechakul, Amit Kohli and Ben Poole for their helpful feedback and discussion. 
This project is supported in part by a Google research scholar award, IARPA DOI/IBC No. 140D0423C0035, and NSF grant No. IIS-2213335. The
views and conclusions contained herein are those of the authors and do not
represent the official policies or endorsements of these institutions.

%
%
\bibliographystyle{plain}
\bibliography{main}

\clearpage
\appendix
\input{sections/X_appendix}
\clearpage

\end{document}

%% file: sections/1_introduction.tex
\section{Introduction}
\label{sec:intro}


Diffusion models have shown tremendous success in modeling complex data distributions like images~\cite{ramesh2022hierarchical,saharia2022photorealistic,balaji2022ediffi,ho2022imagen}, videos~\cite{singer2022make,blattmann2023videoldm} and robot action policies~\cite{chi2023diffusionpolicy}. In domains where data is plentiful, they produce state-of-the-art results. Many data modalities, however, cannot enjoy the same scaling benefits due to their lack of sufficiently large datasets. 
In these cases, it is useful to exploit diffusion models trained on domains with rich data sources as a prior in an optimization framework. 
Score Distillation Sampling (SDS)~\cite{poole2022dreamfusion,Wang_2023_CVPR} and its variants~\cite{wang2023prolificdreamer,hertz2023delta,yu2023textto3d} are a widely adopted way to optimize parametric images, \emph{i.e.}, images produced by a model like NeRF, with a pre-trained diffusion model.
Despite being applicable to a wide range of applications, SDS is also known to suffer from several significant artifacts, such as oversaturation and oversmoothing. 
As such, several variants have been proposed to alleviate these artifacts~\cite{wang2023prolificdreamer,yu2023textto3d,lee2024dreamflow}, often at the cost of efficiency, diversity, or other artifacts.
In this paper, we investigate the core issues with SDS by casting the class of score distillation optimization problems as a Schrödinger Bridge~(SB) problem \cite{Schrödinger1932, Chen2014OnTR,7117355,léonard2013survey}, which finds the optimal transport between two distributions. 
Specifically, given some images from the current optimized distribution (\eg, renderings from a NeRF), applying the transport maps them to their pair images in a target distribution (\eg, text-conditioned natural image distribution).
The density flow formed by these mappings is transport-optimal, as defined in the SB problem.
In an optimization framework, the difference between paired source and target samples, computed with an SB, can be used as a gradient to update the source. 
Su~\etal\cite{su2022dual}
have shown that this path can be explicitly solved using two pre-trained diffusion models. We show that one can also compose these models as an optimizer to approximate transport paths on the fly. 

Under this framework, we can understand SDS and its variants as approximating a source-to-target distribution bridge with the difference of two denoising directions. The denoising scores point to the source and target distributions respectively, with the source representing the current optimized image that updates with each optimization step.

This framing reveals two sources of errors.
First, these methods are a first-order approximation of the diffusion bridge. Specifically, Gaussian noise is sampled to perturb the current optimized image, and single denoising steps, instead of the full PF-ODE simulation, are used to estimate the transport. This induces error in estimating the desired path. Recent works ~\cite{liang2023luciddreamer,lukoianov2024score}
that use multi-step estimation can be explained as mitigating this error. Second, estimating the denoising direction to the current source distribution is non-trivial, since 
the current optimized image may not necessarily look like a real image (\eg, initializing with Gaussian noise or starting from a render of an untextured 3D model). 
Our analysis reveals that SDS approximates the current distribution with the unconditional image distribution, which is not accurate and results in a \textit{distribution mismatch error}. We show that recent SDS variants~\cite{wang2023prolificdreamer,yu2023textto3d,lee2024dreamflow} can be seen as proposals to improve this distribution mismatch error. 

Finally, our analysis motivates a simple method that rectifies the distribution mismatch issue without additional computational overhead. 
Our insight is that the large-scale text-to-image diffusion models learn from billions of caption-image pairs~\cite{schuhmann2022laion}, where a breadth of image corruptions are present in their training sets. 
They are also equipped with powerful pre-trained text encoders, which empower the models with zero-shot capacity in generating unseen concepts~\cite{DALLE,DALLE2}. 
As such, simply describing the current source distribution with text, even if it is not part of the real image manifold, can approximate the distribution of the current optimized image, leading to improved transport paths. 
Our simple and efficient solution can be easily applied to any 
existing application that uses SDS. We show that it consistently improves the visual quality in the desired domain. We comprehensively compare our approach with standard distillation sampling methods over several generation tasks, where our approach matches or outperforms the baselines. 

Our contributions are as follows:

\begin{itemize}
    \item We propose to cast the problem of using a pre-trained diffusion model as a prior in an optimization problem as solving the Schrödinger Bridge (SB) problem between two image distributions. 
    Specifically, it can be seen as bridging the distribution of the current optimized image to the target distribution under a dual-bridge framework. 
    \item We analyze recent SDS-based methods under the lens of our framework and explain the pros and cons of the individual methods.
    \item Our analysis motivates a simple yet effective alternative to SDS by using textual descriptions to specify the current optimized image distribution. 
    It achieves consistently more realistic results than SDS, producing quality comparable with VSD~\cite{wang2023prolificdreamer} without its computational overhead. 
    We compare various generation tasks to show its wall-clock efficiency and quality generations against state-of-the-art methods.
\end{itemize}





%% file: sections/2_related_work.tex
\section{Related Work}

\paragraph{Score Distillation Sampling}
\looseness=-1 Modalities like 3D, 4D, sketch, and vector graphics (SVGs) lack the large-scale, diverse, and high-quality datasets needed to train a domain-specific diffusion model. In these domains, previous works explore exploiting image or video as a proxy modality~\cite{jain2022zero,frans2022clipdraw}. By computing the gradient on a proxy representation with a pretrained model, optimization in the target modality is viable with differentiable mappings, e.g. differentiable rasterization~\cite{li2020differentiable} for SVGs or differentiable rendering~\cite{mildenhall2021nerf} for 3D objects and scenes. The seminal method, Score Distillation Sampling (SDS)~\cite{poole2022dreamfusion}, first proposed to apply a pretrained text-to-image diffusion model for text-to-3D generation. However, it requires a high classifier-free guidance weight and, therefore, suffers from artifacts such as over-saturation and over-smoothing. 
Recent works have built upon SDS to adapt it for editing tasks~\cite{koo2023pds,hertz2023delta,nam2023contrastive,NEURIPS2023_e7fd2c0a} or more broadly improve over the original SDS formulation~\cite{katzir2023noise,alldieck2024score,wang2023prolificdreamer,zhu2023hifa,yu2023textto3d,zou2023sparse3d}. NFSD~\cite{katzir2023noise} and LMC-SDS~\cite{alldieck2024score} inspect the individual components of the SDS gradient and propose methods to rectify the high guidance weights. However, the over-saturation problem is mitigated but not fully resolved. VSD~\cite{wang2023prolificdreamer} formulates the problem as particle-based variational inference and proposes to train a LoRA~\cite{hu2021lora} on the fly to estimate the score of proxy distribution.
We present a new framework that allows rethinking all the variants under the same lens. This framework also motivates a method that improves the quality of SDS without losing efficiency.

\paragraph{Visual Content Generation with SDS}

Since SDS was developed for text-to-3D generation, it has also been adopted to generate various other visual content such as SVGs~\cite{gu2022vector,xing2023svgdreamer}, sketches~\cite{xing2023diffsketcher}, texture~\cite{metzer2023latent,cao2023texfusion,chen2023scenetex,chen2023text2tex,youwang2023paintit}, typography~\cite{iluz2023word}, 3D bodies~\cite{mueller2023buddi}, dynamic 4D scenes~\cite{bahmani20234d,singer2023text,ling2024align} and illusions~\cite{burgert2023diffusion_illusions}.  Among these applications, text-to-3D has been the most active research direction. In addition to designing better distillation sampling methods~\cite{wang2023prolificdreamer,zhu2023hifa,katzir2023noise}, prior work has also studied the underlying 3D neural representations~\cite{yi2023gaussiandreamer,tang2023dreamgaussian,Lin_2023_CVPR,Chen_2023_ICCV} and leveraging multiview data to improve the 3D consistency~\cite{shi2024mvdream,liu2024syncdreamer,Liu_2023_ICCV,qian2024magic,zou2023sparse3d}. We note that these explorations are orthogonal to our study and should be able to work jointly with our method. 
In this paper, we look into existing applications like text-based NeRF optimization, painting-to-real, and illusion generation. We also propose a new AR application called 3D sketch-to-real.

%% file: sections/3_method.tex
\section{Method}
In this section, we present an analytical framework that casts the score distillation sampling (SDS) family of methods as instantiations of a Schrödinger Bridge problem.
We show that many recent SDS based methods can be interpreted as an online solver for the problem. 
That is, each SDS optimization step is a first-order approximation of a dual diffusion bridge formed by two probability flow (PF) ODEs~\cite{su2022dual}. 
We analyze SDS and its variants under this general framework. 
Then, we present a simple solution based on the analysis, which 
leads to significant quality improvement with little extra computational overhead.

\subsection{Background}


\begin{figure}[t]
    \centering
    \includegraphics[width=\linewidth]{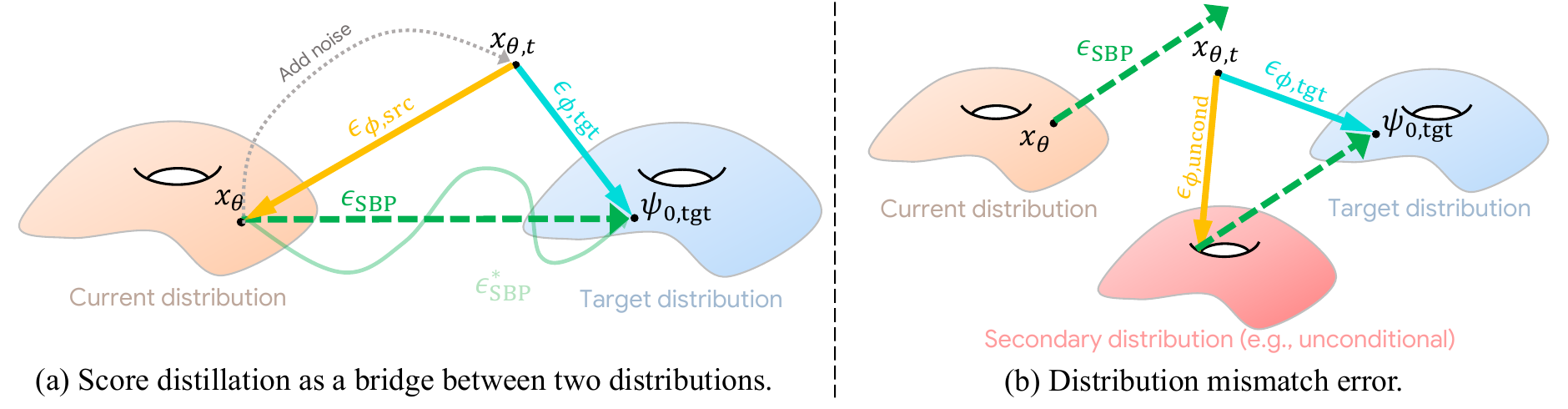}    \caption{\textbf{Optimization with diffusion models as approximation of a Schrödinger Bridge Problem (SBP).} 
    (a) We propose to formulate optimization with diffusion models as bridging the distribution of the current optimized image $x_\theta$ to the target distribution under a dual-bridge framework (a). 
    Current methods can be interpreted as approximating the optimal transport $\epsilon^*_\text{SBP}$ between these distributions via the difference between projections of a noised image $x_{\theta,t}$ onto the two distributions. 
    This analysis reveals two sources of error:
    (1) these gradients are linear approximations of the optimal path, as illustrated in {(a)}, and (2) the source distribution used for computing this approximation (\textit{e.g.}, the unconditional distribution in SDS~\cite{poole2022dreamfusion}) may not be aligned with the current distribution, illustrated in {(b)}.} 
    \vspace{-0.4cm}
    \label{fig:approximate_sb}
\end{figure}

\textbf{Diffusion models} define a forward ``noising" process that degrades data samples $\mathbf{x}$ gradually from the image distribution to noised samples $\mathbf{z}_t$, and eventually the i.i.d. Gaussian distribution~\cite{ho2020denoising,song2021denoising}. 
This process is indexed by timesteps $t$, where $t=1$ indexes the full Gaussian noise distribution and $t=0$ indexes the data distribution. 
A diffusion model, parameterized by $\phi$, is then trained to reverse this encoding process, iteratively transforming the noise distribution into the data distribution with the \cameraready{following denoising} objective:

\begin{equation}
    \label{eq:loss_SDS}
    \mathcal{L}_{\text{Diff}}(\phi, \mathbf{x})=\mathbb{E}_{t \sim \mathcal{U}(0,1), \mathbf{\epsilon} \sim \mathcal{N}(\mathbf{0}, \mathbf{I})}\left[w(t)\left\|\mathbf{\epsilon}_\phi\left(\alpha_t \mathbf{x}+\sigma_t \mathbf{\epsilon} ; y, t\right)-\mathbf{\epsilon}\right\|_2^2\right],
\end{equation}
where \cameraready{$w(t)$ is a loss weighting function}, $y$ is a conditioning text prompt, and $\alpha_t$ and $\sigma_t$ are hyperparameters from the predefined noise schedule. 

\textbf{Probability Flow ODE.}
Denoising score matching~\cite{song2021scorebased,karras2022elucidating,sohl2015deep} shows that the diffusion model denoising prediction can be rewritten as a score vector field: 
\begin{equation}
    \nabla_{\mathbf{x}} \log p_t(\mathbf{x}) = -\frac{1}{\sqrt{1-\alpha}_t} \mathbf{\mathbf{\epsilon}}_t.
    \label{eq:diff_score}
\end{equation}

Because of its special connection to marginal probability densities, the resulting ODE is named the probability flow (PF) ODE with the following expression:

\begin{align}
    dx = [f(\mathbf{x}, t)-\frac{1}{2}g^2(t)\nabla_{\mathbf{x}} \log p_t(\mathbf{x}))] dt,
    \label{eq:pfode}
\end{align}
%
%

%
%
where $f(\mathbf{x}, t)$ and $g(t)$ are pre-defined schedule \cameraready{parameters}. 
This PF-ODE can be solved deterministically \cite{song2022denoising}, mapping a noise sample to its corresponding data sample through the reverse process and the opposite through the forward process (inversion). 
This cycle-consistent conversion between image and latent representations is important in establishing dual diffusion implicit bridges.

\textbf{Dual Diffusion Implicit Bridges.} 
Dual Diffusion Implicit Bridges (DDIBs)~\cite{su2022dual} compose a diffusion inversion and generation process for solving image-to-image translation problems without requiring a paired image dataset. 
Instead, DDIBs use two diffusion models trained on different domains (or, analogously, one model with two different text conditions).  
DDIB inverts the source image into a noise latent via the forward PF-ODE and then decodes the latent in the target domain via the reverse PF-ODE. 
DDIBs can be interpreted as a concatenation of the Schrödinger Bridges from source-to-latent and latent-to-target, hence the dual bridges in its name. 
DDIBs enable solving transport between two distributions using a single pre-trained diffusion model. 
We build on this insight in an optimization context. 

\subsection{Optimization with Diffusion Model Approximates a Dual Schrödinger Bridge}

Many generative vision tasks involve optimizing corrupted images to the image manifold. 
For example, in 3D generation, a 3D representation like NeRF is optimized to render natural images matching a prescribed text prompt. 
Methods like SDS enable this by using a pre-trained diffusion model as a prior.
We propose formulating such optimization problems as solutions to an instantiation of a Schrödinger Bridges Problem (SBP).
SBP finds cost-optimal paths between a source image distribution $p_\text{src}$ and a target image distribution $p_\text{tgt}$~\cite{wang2021deep,de2021diffusion}. 
Optimizing a parametrized image toward the natural image distribution can be cast as finding the optimal paths between the current optimized image(s) and the natural image distribution.
Instead of solving this problem directly, which would require training a generative model from scratch ~\cite{liu2023i2sb,de2021diffusion,chen2022likelihood}, we show that pre-trained diffusion models can be exploited as an optimizer that approximates the path. Further, the gradient computed by the existing score distillation methods can be viewed as the first-order approximation of this path. This formulation is illustrated in Figure~\ref{fig:approximate_sb}.

Let $\mathbf{x}_\theta\in \mathbb{R}^d$ represent a parametric image, \ie, an image produced differentiably by a model with parameter $\theta$, such as a NeRF. 
To leverage the pretrained diffusion model, we add noise $ \mathbf{\epsilon} \sim \mathcal{N}(\mathbf{0}, \mathbf{I})$ to obtain a latent at timestep $t$:

\begin{equation}
    \mathbf{x}_{\theta, t}=\alpha_t \cameraready{\mathbf{x}_{\theta}}+\sigma_t \mathbf{\epsilon}
\end{equation}

Suppose that $\psi_{t^\prime, \text{src}}$ and $\psi_{t^\prime, \text{tgt}}$ denote the paths obtained by solving the PF ODE as in Eq. \ref{eq:pfode} from $t$ to $0$, both starting from  $\mathbf{x}_{\theta, t}$, such that $\psi_{0, \text{src}} \in p_\text{src}$, $\psi_{0, \text{tgt}} \in p_\text{tgt}$, $\psi_{t, \text{src}}=\psi_{t, \text{tgt}}=\mathbf{x}_{\theta, t}$. 
This forms a dual diffusion bridge \cite{su2022dual} from $\psi_{0, \text{src}}$ to $\psi_{0, \text{tgt}}$. 
We approximate this path \textit{per-iteration} using a pretrained diffusion model. 
We denote the displacement of this path as:

\begin{equation}
    \mathbf{\epsilon}_\text{SBP}^* = \psi_{0, \text{tgt}} - \psi_{0, \text{src}}.
\end{equation}

Fully simulating this bridge involves solving two PF ODEs, which invokes dozens of neural function evaluations (NFEs) to estimate the gradient of each iteration. Instead, one can estimate each half of the bridge with a single-step prediction
by computing two denoising directions $\mathbf{\epsilon}_{\phi, \text{src}}$ and $\mathbf{\epsilon}_{\phi, \text{tgt}}$. 
We thus obtain a first-order approximation of a dual diffusion bridge with the difference vector:
\begin{equation}
    \mathbf{\epsilon}_\text{SBP} = \mathbf{\epsilon}_{\phi, \text{tgt}}-\mathbf{\epsilon}_{\phi, \text{src}},
\end{equation}
which is subject to the following sources of errors.

\begin{enumerate}
    \item \textbf{First-order approximation error}. Instead of performing full PF-ODE simulations, the single-step noising and prediction are less accurate and induce errors. Recent work ISM~\cite{liang2023luciddreamer} can be interpreted as reducing this error with a multi-step simulation to obtain $\mathbf{x}_{\theta, t}$.
    \item \textbf{Source distribution mismatch}. The dual diffusion bridge relies on $\mathbf{\epsilon}_{\phi, \text{src}}$ accurately estimating the distribution of the current sample, $\mathbf{x}_\theta$. A series of works can be viewed as improving this error~\cite{wang2023prolificdreamer,katzir2023noise,yu2023textto3d} by computing more accurate $\mathbf{\epsilon}_{\phi, \text{src}}$ .
\end{enumerate}



We show that  $\mathbf{\epsilon}_{\phi, \text{tgt}}-\mathbf{\epsilon}_{\phi, \text{src}}$ is an effective gradient when both the source and target distribution are well expressed. Next, we discuss the popular score distillation methods under this analysis. We argue that their characteristic artifacts can largely be understood due to the errors above.

\begin{figure}[t]
    \centering
    \includegraphics[width=\linewidth]{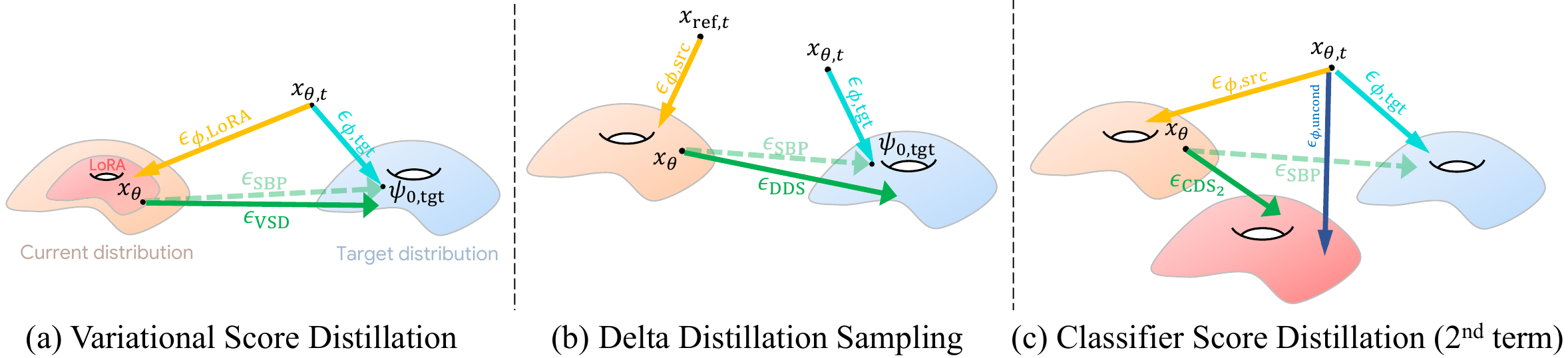}
    \caption{\textbf{Comparision of SDS variants under our analysis}. 
    We illustrate the major gradient components of different SDS variants and provide a straightforward comparison with $\mathbf{\epsilon}_\text{SBP}$. }
    \vspace{-0.4cm}
    \label{fig:sds_analysis}
\end{figure}

\subsection{Analyzing Existing Score Distillation Methods}

We analyze SDS and its variants through our framework by inspecting each component in the computed gradient. 
For notation, $y_\text{tgt}$ is the text prompt representing the target distribution, and $\varnothing$ denotes the unconditional prompt. 
For each method, we present its gradient update and discuss its implications.


 \topic{Score Distillation Sampling~\cite{poole2022dreamfusion}:} 
 $$\mathbf{\epsilon}_\text{SDS} = \mathbf{\epsilon}_\phi\left(\mathbf{x}_{\theta, t}; \varnothing, t\right) + s\cdot(\mathbf{\epsilon}_\phi\left(\mathbf{x}_{\theta, t}; y_\text{tgt}, t\right) - \mathbf{\epsilon}_\phi\left(\mathbf{x}_{\theta, t}; \varnothing, t\right)) - \mathbf{\epsilon},$$
where $s$ is the strength of classifier-free guidance. 
When $s$ is small, the $\mathbf{\epsilon}$ functions as an averaging term to regress the image to the mean. 
However, the SDS gradient has been shown to work best with extreme values of classifier-free guidance $s$ like $100$. 
We can rewrite the gradient to emphasize how the conditional-unconditional delta dominates at high CFG scales.

$$\mathbf{\epsilon}_\text{SDS} = \underbrace{s\cdot(\mathbf{\epsilon}_\phi\left(\mathbf{x}_{\theta, t}; y_\text{tgt}, t\right)-\mathbf{\epsilon}_\phi\left(\mathbf{x}_{\theta, t}; \varnothing, t\right))}_{\text{Dominant when } s \gg 1} + \mathbf{\epsilon}_\phi\left(\mathbf{x}_{\theta, t}; \varnothing, t\right) - \epsilon,$$

Experimentally, we produce very similar results at high CFG with or without the non-dominant terms. 
We argue that SDS should be interpreted through the dominant term, which fits within our analysis. 
Under this interpretation, \cameraready{the unconditional direction $ \phi\left(\mathbf{x}_{\theta, t}; \varnothing, t\right)$ approximates the source distribution of $\textbf{x}_\theta$ poorly, instead representing images of any identity with low contrast and geometric artifacts. }
Figure~\ref{fig:approximate_sb}(b) illustrates the effect of a poor approximation. 
The bridge from the unconditional to conditional distribution leads to the characteristic oversaturation and smoothing of SDS results. 

\topic{Delta Distillation Sampling~\cite{hertz2023delta}:} 
$$\mathbf{\epsilon}_\text{DDS} = \mathbf{\epsilon}_\phi\left(\mathbf{x}_{\theta, t}; y_\text{tgt}, t\right)-\mathbf{\epsilon}_\phi\left(\mathbf{x}_{\text{ref}, t}; y_\text{src}, t\right), $$
where $\mathbf{x}_{\text{ref}, t}$ is a noised version of a reference image in the image editing task. 
As shown in Figure~\ref{fig:sds_analysis} (b), this increases the \textit{source distribution mismatch} since $\mathbf{\epsilon}_{\phi, \text{src}}$ is not calculated based on the current optimized image $\mathbf{x}_{\theta, t}$.

\topic{Noise Free Score Distillation~\cite{katzir2023noise}:}
 $$\mathbf{\epsilon}_\text{NFSD} = (\mathbf{\epsilon}_\phi\left(\mathbf{x}_{\theta, t}; \varnothing, t\right)- (t<0.2)\cdot  \mathbf{\epsilon}_\phi\left(\mathbf{x}_{\theta, t}; y_\text{neg}, t\right)) + s\cdot(\mathbf{\epsilon}_\phi\left(\mathbf{x}_{\theta, t}; y_\text{tgt}, t\right) - \mathbf{\epsilon}_\phi\left(\mathbf{x}_{\theta, t}; \varnothing, t\right)),$$
where the strength of classifier-free guidance $s$ is set to $7.5$ and $y_\text{neg}=$``unrealistic, blurry, low quality ...''. NFSD greatly reduces the guidance strength while it is observed to perform very similarly to SDS in practice. We can better explain this phenomenon since the prompt $y_\text{neg}$ does not accurately describe the source distribution as it omits the image's content. 
In addition, the second component with weight $s=7.5$ still forms the major part of the gradient, which is the dominant term in SDS.


\topic{Classifier Score Distillation~\cite{yu2023textto3d}:} 
$$\mathbf{\epsilon}_\text{CSD} = w_1\cdot (\mathbf{\epsilon}_\phi\left(\mathbf{x}_{\theta, t}; y_\text{tgt}, t\right) - \mathbf{\epsilon}_\phi\left(\mathbf{x}_{\theta, t}; \varnothing, t\right)) + w_2\cdot (\mathbf{\epsilon}_\phi\left(\mathbf{x}_{\theta, t}; \varnothing, t\right) - \mathbf{\epsilon}_\phi\left(\mathbf{x}_{\theta, t}; y_\text{src}, t\right)), $$
where $w_1$ and $w_2$ are hyperparameters. As shown in Figure~\ref{fig:sds_analysis} (c), the second term approximates the bridge from the source distribution to the unconditional distribution, which is not ideal since it does not point to the target distribution. It explains the observation made by the authors~\cite{yu2023textto3d} that this undermines the alignment with the text prompt. Therefore, the authors always anneal $w_2$ to $0$ during the optimization. 
However, we show this often reintroduces the SDS artifacts in practice.

\topic{Variational Score Distillation~\cite{wang2023prolificdreamer,lee2024dreamflow}:} 
$$\mathbf{\epsilon}_\text{VSD} =\mathbf{\epsilon}_\phi\left(\mathbf{x}_{\theta, t}; \varnothing, t\right) + s\cdot(\mathbf{\epsilon}_\phi\left(\mathbf{x}_{\theta, t}; y_\text{tgt}, t\right) - \mathbf{\epsilon}_\phi\left(\mathbf{x}_{\theta, t}; \varnothing, t\right))-\mathbf{\epsilon}_{LoRA}\left(\mathbf{x}_{\theta, t}; y_\text{tgt}, t\right).$$

Out of all the discussed methods, VSD attempts to minimize the \textit{source distribution mismatch} error most directly by test-time finetuning a copy of the diffusion model with LoRA on the current set of $\mathbf{x}_\theta$. 
Note that in the original paper, the use of LoRA was motivated based on a particle-based variational framework. Our analysis enables an alternative understanding of VSD. 
As shown in Figure~\ref{fig:sds_analysis} a), this approach is well-justified in our dual diffusion bridge framework. 
However, training a LoRA \emph{every iteration} is computationally expensive, adds complexity, and introduces its own low-rank approximation errors. 
Given this insight, we propose a simple yet efficient approach to mitigating source distribution without LoRA. 


\subsection{Mitigating Source Distribution Mismatch with Textual Descriptions}

Our analysis reveals that the LoRA model in VSD most closely approximates the distribution of the current optimized parametrized image, addressing the distribution mismatch error. 
Unfortunately, it incurs $200-300\%$ runtime overhead on top of SDS, making it impractical, despite its significant performance gains.
With this understanding, we propose a simple approach that better expresses the source distribution. 
Our insight is that pre-trained diffusion models have learned the distribution of natural and corrupted images through a combination of powerful text representation and enormous image-caption datasets. 
We find that by simply describing image corruptions with a text prompt, we can improve our estimate of the source distribution.

Specifically, we propose to use the gradient

$$\mathbf{\epsilon}_\text{ours}=w\cdot(\mathbf{\epsilon}_\phi\left(\mathbf{x}_{\theta, t}; y_\text{tgt}, t\right) - \mathbf{\epsilon}_\phi\left(\mathbf{x}_{\theta, t}; y_\text{src}, t\right)),$$

where we get $y_\text{src}$ by adding descriptions of the current image distribution to $y_\text{tgt}$ (the base prompt). The remaining question is how to set this description. In generation tasks, we propose a simple two-stage solution.

\begin{enumerate}
    \item We use $\mathbf{\epsilon}_\text{SDS}$ to produce a generation with the method's characteristic artifacts:
    \item We switch to optimization with our gradient, $\mathbf{\epsilon}_\text{ours}$, to transport the image parameter toward the natural image distribution.
\end{enumerate}

To describe the artifacts produced by SDS, we append the descriptors ``, \texttt{oversaturated, smooth, pixelated, cartoon, foggy, hazy, blurry, bad structure, noisy, malformed}'' and drop the descriptors of the high-quality generation. 
\cameraready{This description $y_\text{src}$ does not require hand-crafting based on problem domains—it is fixed across all shown examples and use cases. As shown in Appendix Figure~\ref{fig:ablate_prompt}, we explored searching for other prompts but did not find that variations in these descriptions made a big difference.}


In editing tasks, we have an initialization that $y_\text{src}$ describes accurately. In such cases, we omit the first SDS stage and only apply our gradient to optimization. We also append a ``domain descriptor.'' For instance, in painting-to-real, this is simply ``, painting'' to represent the initial distribution.

While the use of such negative prompting has been explored before, such as in NFSD, our analysis motivates a principled way to incorporate it into score distillation. We find that these simple modifications significantly narrow the quality gap between SDS and resource-intensive methods like VSD. We verify this finding experimentally with qualitative results and quantitative comparisons across applicable tasks.

%% file: sections/4_experiments.tex
\section{Experiments}

\looseness=-1 
In this section, we test our proposed method on several generation problems where SDS is adopted. 
We compare against SDS and other task-specific baselines. 
Note that our goal is not to show another state-of-the-art text-to-3D generation method, but to verify our findings, where the proposed score distillation approach based on textual description efficiently improves the results by mitigating the source distribution mismatch error. 
We first perform a thorough experiment in a controlled setting on text-to-image generation. 
Then, we compare it on text-guided NeRF optimization to SDS and VSD and evaluate the painting-to-real image translation task against image editing baselines. 
\cameraready{Please see more results in the appendix, including additional qualitative results and comparison, ablation studies and our method's application to optical illusion generation and 3D-sketch-to-real task.} 


\begin{figure}[t]
    \centering
    \includegraphics[width=\textwidth]{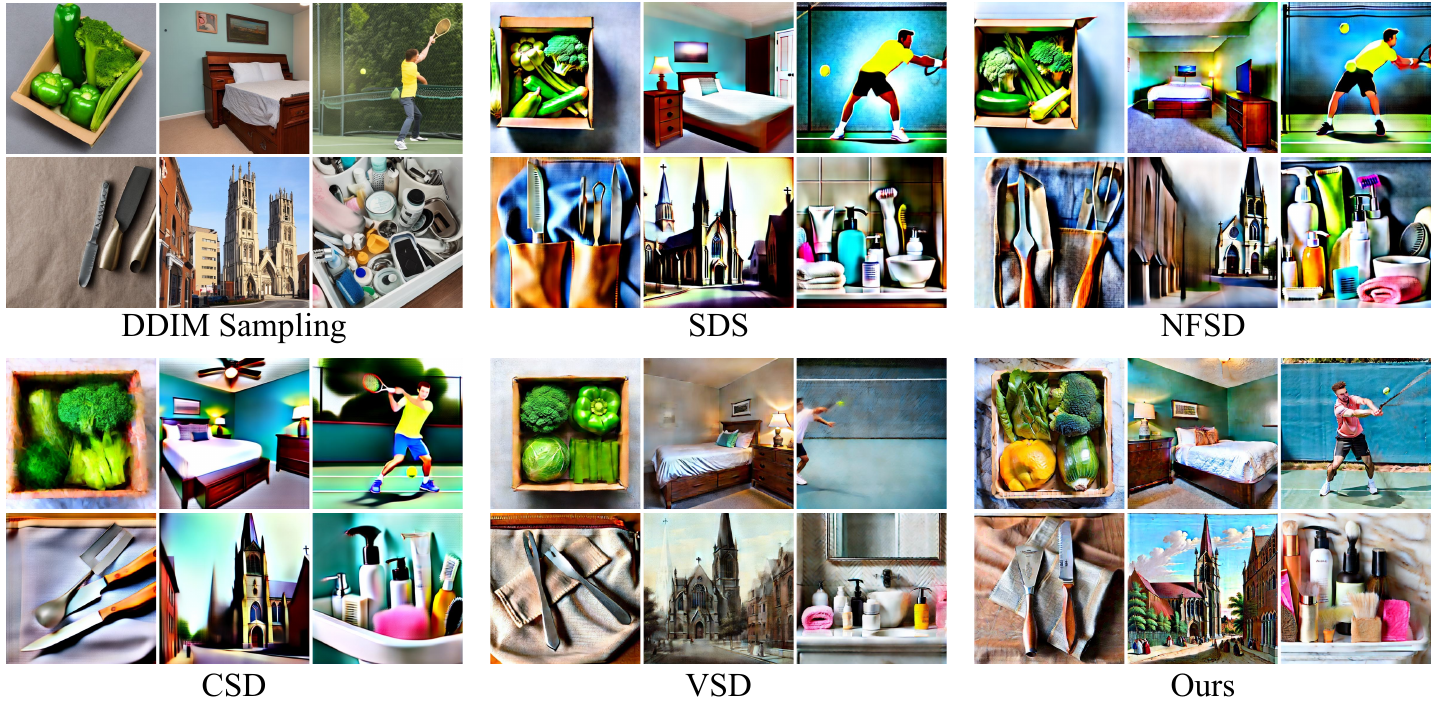}
    \caption{\textbf{Text-to-image generation results with COCO Captions.} We compare different score distillation methods for generating images with COCO captions by optimizing a randomly initialized image. DDIM sampling indicates the lower bound that the diffusion model can achieve. VSD~\cite{wang2023prolificdreamer} and our method generate the least color artifacts while ours is more efficient than VSD.}
    \vspace{-0.4cm}
    \label{fig:coco_visual}
\end{figure}

\begin{table}[t]
\setlength{\tabcolsep}{5.5pt}
\small
\caption{\textbf{Zero-shot FID comparison with different score distillation methods.} We report FID scores of text-to-image generation using 5K captions randomly sampled from the COCO dataset. The best score distillation result is indicated in \textbf{bold}, while the second best is \underline{underlined}.}
\begin{tabular}{@{}lllllll@{}}
\toprule
  & DDIM (\footnotesize{lower bound}) & SDS~\cite{poole2022dreamfusion} & NFSD~\cite{katzir2023noise} & CSD~\cite{yu2023textto3d} &  VSD~\cite{wang2023prolificdreamer}  & Ours \\ \midrule
Zero-Shot FID ($\downarrow$) & $49.12$ & $86.02$ & $91.70$ & $89.96$ & $\mathbf{59.22}$ & \underline{$67.89$} \\
Zero-Shot CLIP FID ($\downarrow$)& $16.56$ & $28.39$ & $29.25$ & $27.07$ & $\mathbf{18.86}$ & \underline{$20.31$} \\
Time per Sample (mins) & $0.05$ & $\mathbf{4.48}$ & $7.20$ & \underline{$6.21$} & $16.02$ &         $\mathbf{4.48}$  \\ \bottomrule
\end{tabular}
\vspace{-0.4cm}
\label{tab:coco_fid}
\end{table}

\subsection{Zero-Shot Text-to-Image Generation with Score Distillation} 

To verify our analysis of existing SDS variants and the proposed method, we perform text-to-image generation by optimizing an image of size $64\times64\times4$ in the Stable Diffusion latent space~\cite{wang2023prolificdreamer,katzir2023noise} (\cameraready{We explore other base models like MVDream~\cite{shi2024mvdream} and SDXL~\cite{podell2023sdxl} in Appendix Figure~\ref{fig:mvdreamsdxl}}). 
The benefit of choosing image generation as the evaluation task is that its generation quality has the least confounding variables among other tasks. (\emph{e.g.,} in text-to-3D, many designs like regularizations~\cite{zhu2023hifa}, initialization~\cite{Lin_2023_CVPR}, 3D representations~\cite{Chen_2023_ICCV,tsalicoglou2023textmesh,yi2023gaussiandreamer,tang2023dreamgaussian}, and 2D prior models~\cite{shi2024mvdream,liu2024syncdreamer,Liu_2023_ICCV,qian2024magic,zou2023sparse3d} could affect the final quality.)

We use the MS-COCO~\cite{lin2014microsoft} dataset for the evaluation. 
Consistent with the prior study~\cite{balaji2022ediffi}, we randomly sample 5K captions from the COCO validation set as conditions for generating images. 
For each caption, we optimize a randomly initialized the image with the score distillation gradients. 
We compare our method with several SDS variants including SDS~\cite{poole2022dreamfusion}, NFSD~\cite{katzir2023noise}, CSD~\cite{yu2023textto3d}, and VSD~\cite{wang2023prolificdreamer}. 
For all the methods, we use the same learning rate of $0.01$ and optimize for $2,500$ steps where we generally observe convergence. 
We compute the zero-shot FID~\cite{heusel2017gans} and CLIP FID scores~\cite{kynkaanniemi2022role} between these generated images and the ground truth images. We also report results generated by DDIM with $20$ steps as a lower bound for renference.

\looseness=-1 We report the FID scores and the time to optimize one image in Table~\ref{tab:coco_fid}. 
Among all the score distillation methods, VSD~\cite{wang2023prolificdreamer} achieves the lowest FID scores. 
However, it requires training a LoRA along the optimization process. 
Instead, ours achieves a comparable FID score with over $3\times$ faster speed. 
We visualize random examples generated by different score distillation methods in Figure~\ref{fig:coco_visual}. 
We notice that SDS and NSFD suffer from the over-saturation and over-smoothness issues. 
CDS has slightly fewer color artifacts. VSD and ours generate the samples that most closely resemble the DDIM sampling.


\begin{figure}[t!]
    \centering
    \includegraphics[width=\linewidth]{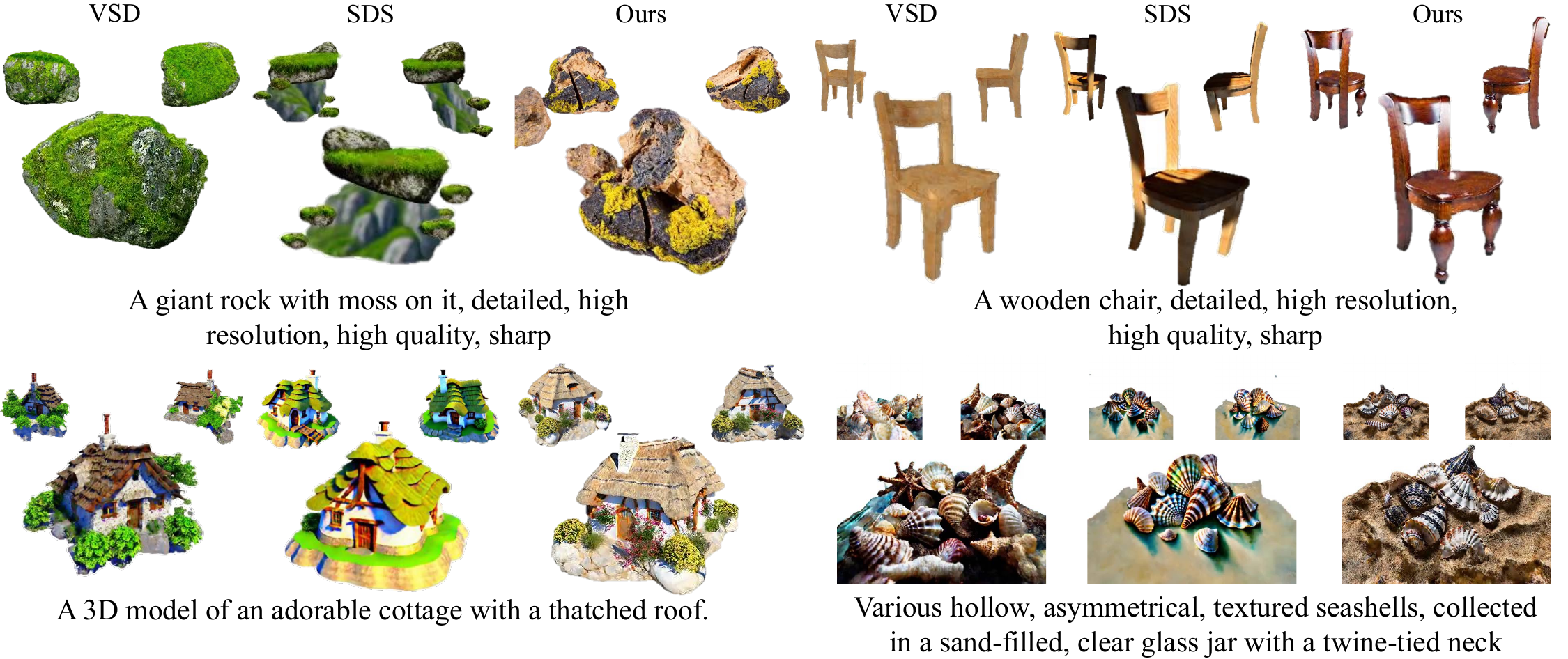}
    \caption{\textbf{Text-guided NeRF optimization with different score distillation methods.} 
    We make a fair comparison of SDS and VSD for text-to-3D generation. 
    For each generation, we show three uniformly sampled views. 
    SDS results like the cottage and pepper mill still suffer from over-saturation problems, while ours and VSD can produce realistic details, color, and texture. }
    \label{fig:text_to_3d}
    \vspace{-0.4cm}
\end{figure}

\subsection{Text-guided NeRF Optimization} 

We now evaluate the text-to-3D generation problem, where we intentionally aim to exclude variables that could affect the generation quality other than the score distillation methods. We use the ThreeStudio~\cite{threestudio2023} repository to optimize a NeRF with settings tuned for ProlificDreamer stage 1 (NeRF optimization)~\cite{wang2023prolificdreamer}. Note that we do not perform stages 2 and 3, \ie geometry fine-tuning and texture refinement. Specifically, we initialize the NeRF with the method proposed by Magic3D~\cite{Lin_2023_CVPR}, use the regularization losses on the sparsity and opacity, and optimize for 25K steps. We adopt the native SDS and VSD guidance implementations for comparison. \cameraready{In Appendix Figure~\ref{fig:more3d}, we evaluate our methods with additional text-to-3D systems, including Fantasia3D~\cite{Chen_2023_ICCV}, Magic3D~\cite{Lin_2023_CVPR} and CSD~\cite{yu2023textto3d}.}

\begin{wrapfigure}[10]{r}{0.45\textwidth}
\centering
\footnotesize
\captionof{table}{\textbf{Quantitative comparisons of NeRF optimization}.
We measure the average CLIP similarity of rendered views using SDS, VSD
and our.}
\label{tab:text_to_3d_clip}
\begin{tabular}{@{}llll@{}}
\toprule
     & ViT-L/14 & ViT-B/16 & ViT-B/32 \\ \midrule
SDS~\cite{poole2022dreamfusion}  & 0.2811   & 0.3196   & 0.3139   \\
VSD~\cite{wang2023prolificdreamer}  & 0.2837   & 0.3292   & 0.3166   \\
Ours & 0.2848   & 0.3282   & 0.3148   \\ \bottomrule
\end{tabular}
\end{wrapfigure}

We first show visual comparisons of different score distillation methods in Figure~\ref{fig:text_to_3d}. 
We notice that SDS tends to generate fewer details, as shown by the rock and chair examples, and sometimes suffers from over-saturation issues, as in 2D, as demonstrated by the cottage and seashell examples. Instead, both VSD and ours can generate highly photo-realistic 3D objects, while ours does not require training a LoRA model and shares a similar computational cost as SDS.  


We also perform a quantitative evaluation and user study on the NeRFs optimized based on 31 different text prompts. Note that this number is similar to the choice of existing works on the text-to-3D task~\cite{liang2023luciddreamer,lee2024dreamflow,dong2024coin3d}. 
However, different from these works that ignore the confounding 3D variables that contribute to the generation quality, we disentangle this by isolating the score distillation method as the only comparison variable. We follow these works to evaluate the generation quality with CLIP~\cite{radford2021learning}. 
We report the CLIP similarity in Table~\ref{tab:text_to_3d_clip}. 
Our method consistently outperforms SDS and achieves comparable results with VSD. In addition, in a user study consisting of 37 users, shown pairwise comparisons of rotating 3D renders (\emph{i.e.,} comparisons of our result and a random choice of VSD or SDS, with the prompt: ``For a text-to-3D system, given the prompt \textit{[p]}, which result would you be happiest with?''), our results were chosen in 75.7\% of all responses. 




\begin{figure}[t!]
    \centering\includegraphics[width=\linewidth]{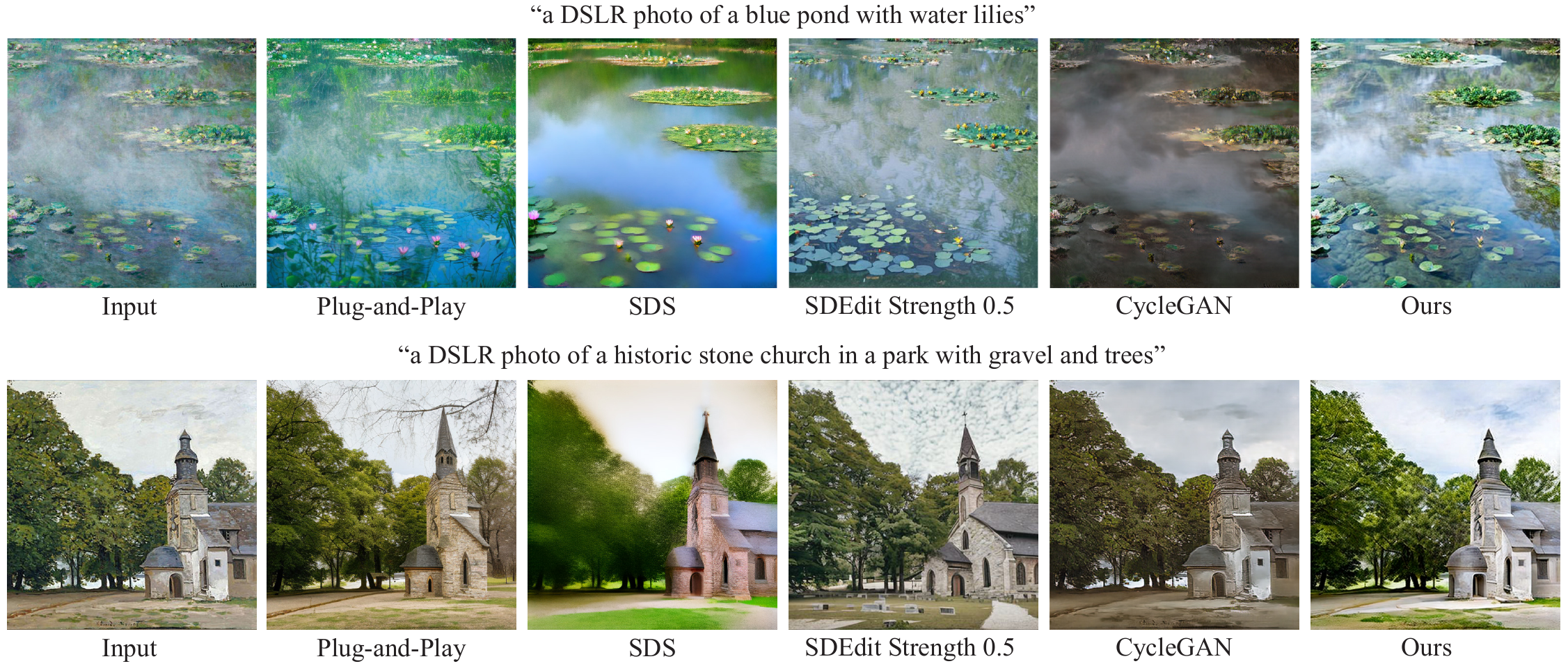} 
    \caption{\textbf{Painting-to-Real comparison.} We compare our gradient in optimization to image restoration and image-conditional generation baselines. While SDEdit produces convincing textures, it is difficult to find a strength value that balances structure and quality. Other baselines fail to reproduce natural image quality, while our method produces the best combination of quality and faithfulness.}
    \vspace{-0.2cm}
    \label{fig:paint2real_comparison}
\end{figure}


\subsection{Painting-to-Real}
We examine our method's ability to serve as a general-purpose realism prior. Paintings are "near-manifold" images, meaning they do not possess natural image statistics but live near the image distribution in image space. An effective image prior 
should guide a painting toward a nearby natural image through optimization.

We initialize a latent image by encoding scans of the artwork through Stable Diffusion's encoder. We specify a prompt for each painting to condition the diffusion model and then apply the second optimization stage of our method (SDS stage omitted). We experimented with automatically generating prompts via pretrained vision language models but found the results inconsistent, so we leave this to future work. 
Since the large image datasets used to train diffusion models contain artwork, we append the domain descriptor ``, painting'' to $y_\text{src}$ to optimize away from this distribution.

\begin{figure}[!ht]
    \centering
    \includegraphics[width=\linewidth]{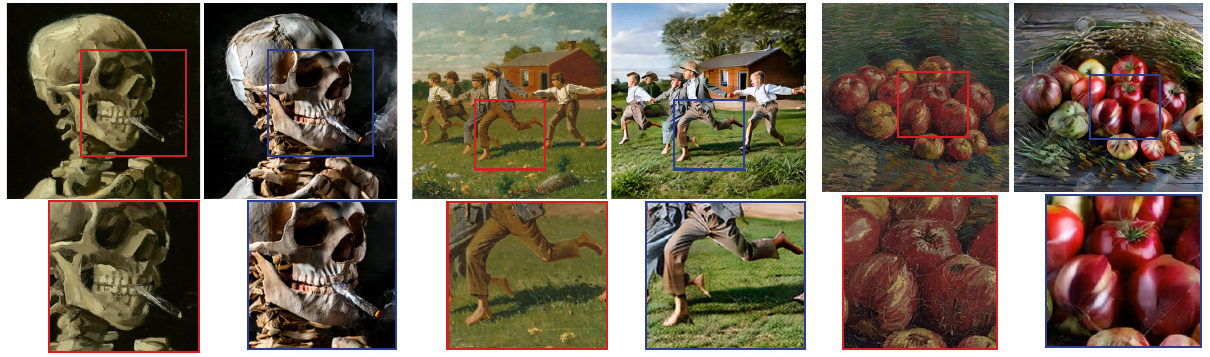}
    \caption{\textbf{Painting-to-Real results.}\looseness=-1  We show selected Painting-to-Real samples with diverse art styles and subjects. Initialization images are shown on the left, optimized images are shown on the right.}
    \vspace{-0.2cm}
    \label{fig:paintingtorealgallery}
\end{figure}

While SDS is proposed to leverage a pretrained text-to-image diffusion model as an image prior, its artifacts make it ineffective in practice. 
In comparison, our method realistically synthesizes details and relights the image naturally. We observe that SDS methods diverge more easily in 2D experiments than in 3D but that the issue can be mostly resolved with tuning. A future goal is to formulate a gradient that can be applied idempotently \cite{shocher2024idempotent}. We compare with image reconstruction baselines in Figure~\ref{fig:paint2real_comparison} and provide a small gallery of painting-to-real results in Figure~\ref{fig:paintingtorealgallery}.




%% file: sections/4.5_discussion.tex
\section{\cameraready{Discussion on Solving the Linear Approximation Error}}

\begin{table}[t]
\setlength{\tabcolsep}{4pt}
\caption{\cameraready{\textbf{Reducing first-order approximation error improves generation quality.}  Using full PF-ODE simulation (``Full-path'') to replace single-step prediction improves visual quality in all settings.}}
\label{tab:full_path}
\begin{tabular}{l|lll|lll}
\toprule
Approximate the bridge & \multicolumn{3}{c|}{Single-step}         & \multicolumn{3}{c}{Full-path} \\
Estimate source distribution   & Uncond. (SDS) & Bridge & LoRA (VSD) & Uncond.     & Bridge & LoRA  \\
\midrule
Zero-Shot COCO FID ($\downarrow$)     & 86.02    & 67.89  & 59.22      & 63.31  & 60.07  & 55.65 \\
\bottomrule
\end{tabular}
\vspace{-0.3cm}
\end{table}

\cameraready{As we have shown that reducing the distribution mismatching error can significantly improve the generation quality of the score distillation optimization, it is natural to ask whether one can also reduce the first-order approximation error, induced by linear bridge estimation, to improve the results further. Several recent studies, including SDI~\cite{lukoianov2024score} and ISM~\cite{liang2023luciddreamer}, can be viewed as mitigating this error by replacing the single-step estimation with a multi-step estimation to an intermediate timestep. 
Under our framework, one can estimate the entire dual bridge by solving both PF-ODE paths. Specifically, via inversion, one can solve the PF-ODE path from $\psi_{0, \text{src}}$ to $x_{\theta, T}$, and then walk to the $\psi_{0, \text{tgt}}$ via sampling.
In this way, it is possible to obtain the most accurate gradient direction with little approximation error $\mathbf{\epsilon}_\text{SBP}^*=w\cdot(\psi_{0, \text{tgt}} -\psi_{0, \text{src}})$. We refer to this approach as ``full path''. Note that this resolves the linear approximation error, and it is independent of handling the source approximation error, which could be addressed via the discussed text description or LoRA.}

\cameraready{However, solving the inversion ODE is not trivial~\cite{karras2022elucidating}. We noticed that the inversion can exaggerate the distribution mismatch error and cause the optimization to get stuck at a local optimum at the beginning of the optimization. Instead, the stochasticity of the single-step methods often shows more robustness to the input image. Therefore, we first perform the single-step score distillation optimization to obtain reasonable results and then switch to solving the full bridge. We also anneal the timestep endpoint of the bridge throughout the optimization. With this approach, we can now explore addressing both the first and second sources of error. The first source (linear approximation) has ``full-path,'' and the second source (source distribution mismatch error) has ``Bridge'' or ``LoRA''. We find that using the ``full-path'' multi-step  (mitigating linear approximation error)
always outperforms the single-step methods, achieving a lower FID, as shown in Table~\ref{tab:full_path}.
However, the same trend does not fully transfer to the text-to-3D experiments. We observe that solving the entire bridge typically introduces additional artifacts and makes the optimization less stable. We leave the best way of leveraging this gradient for future research exploration.}

%% file: sections/5_conclusion.tex
\section{Conclusion}
We present an analysis that formulates the use of a pre-trained diffusion model in an optimization framework as seeking an optimal transport between two distributions. Under this lens, we analyze SDS variants with a unified framework. We also develop a simple approach based on textual descriptions that work comparably well to the best-performing approach, VSD, without its significant computational burden. However, neither approach has yet to achieve the quality and diversity of images generated by the reverse process. We hope that our analysis enables the development of a more sophisticated solution that can one day achieve the same quality and diversity as the reverse process in an optimization framework. Combining our proposed method with multi-step approximations like ISM~\cite{liang2023luciddreamer} or schedules like DreamFlow~\cite{lee2024dreamflow} could mitigate the first-order approximation error and further improve the efficiency, which is an interesting future research direction.  With the rise of high-quality video diffusion models, we anticipate that the question of how to effectively use such models as a prior in various problems will become even more important.

%% file: sections/X_appendix.tex
\textbf{\Large{Appendix}}\\

In this appendix, we discuss the additional experiment details and provide more visual results, including optical illusion sketch, text-based NeRF optimization, and 3D paint-to-real results. We also perform an ablation study of our method.

\section{Additional Experimental Setup}
In this section, we describe our experimental setups in more detail. 

\paragraph{Text-to-image generation with score distillation.} \cameraready{We use the \textit{stable-diffusion-v2-1-base} model by default for our experiments if not specified.} For CSD, we follow the original paper~\cite{yu2023textto3d} to use $w_1=w_2=40$ at the initialization steps and anneal $w_2=0$ within the first $500$ steps. We use $s=100$ for SDS and $s=7.5$ for NFSD and VSD, which are consistent with the best practice. We use $s=40$ and $w=25$ for our method. And we optimize with $\mathbf{\epsilon}_\text{SDS}$ loss for $500$ iterations and then switch to $\mathbf{\epsilon}_\text{ours}$ for the rest of $2,000$ iterations.  For all the methods, we use a learning rate of $0.01$, and we use a learning rate of $1e-4$ to train the LoRA in VSD.

\paragraph{Text-guided NeRF optimization with score distillation.} For our method, we optimize with $\mathbf{\epsilon}_\text{SDS}$ loss for $20,000$ iterations and then switch to $\mathbf{\epsilon}_\text{ours}$ for the rest of $5,000$ iterations. We use $s=100$ and $w=1$ for our method. We find that a high $s$ is necessary to establish geometry in the first stage of the text-to-3D setting, but our method is not too sensitive to this hyperparameter in 2D. We use the rest of the learning rates and regularization strengths as the default settings. 


\section{More Visual Results}
\renewcommand{\thefigure}{A\arabic{figure}}
\setcounter{figure}{0}
In this section, we provide extra visual results. Specifically, we show 3D sketch-to-real and optical illusion generation as additional applications of our method. We also report more comparisons and ablation studies of text-based NeRF optimzition.

\begin{figure}[t!]
    \centering
    \includegraphics[width=\linewidth]{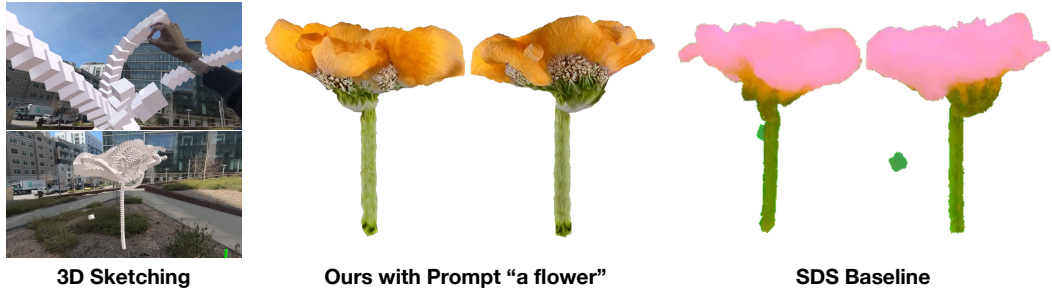}
    \caption{\textbf{3D sketch-to-real.} We introduce a conditional generation task in 3D where a coarse human-drawn mesh is optimized into a high-quality mesh. While SDS and our gradient both adhere to the prompt and shape conditions, our method produces higher fidelity colors and texture.}
    \vspace{-0.4cm}
    \label{fig:sketch_to_real_comparison}
\end{figure}

\subsection{Additional Applications}

\paragraph{3D Sketch-to-Real}
Head-mounted displays with hand tracking are a natural platform for a sort of "3D sketching," where 3D primitives trail from your hand like ink from a pen. The resulting coarse mesh is structurally accurate but lacks geometric or texture detail. To this end, we propose a new application that transfers these 3D sketches to more realistic versions. We extend our text-to-3D solution to generate these details.

We first fit an implicit SDF volume to multi-view renders of the mesh, then apply our gradient with the same schedule as in text-based NeRF optimization. We lower the learning rate for geometry parameters to prevent divergence from the guiding sketch. Holding other hyperparameters equal, we compare our gradient and the SDS gradient in Figure \ref{fig:sketch_to_real_comparison}.

\begin{figure}[t]
    \centering
    \includegraphics[width=\linewidth,trim=0 0 0 0,clip]{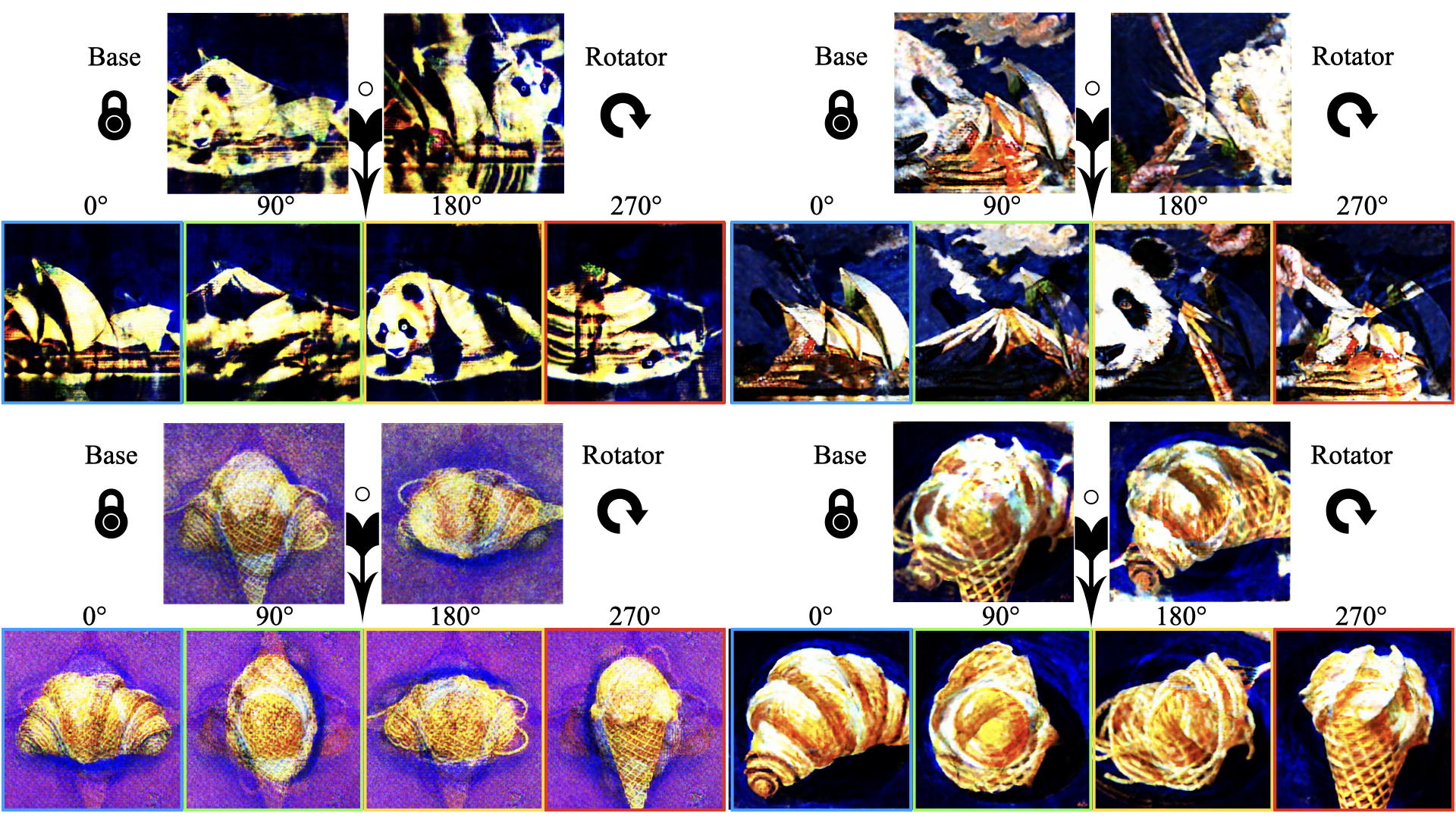}
    \begin{minipage}[c]{0.45\linewidth}
        \hspace{0.4\linewidth}
         SDS~\cite{poole2022dreamfusion}
    \end{minipage}
    \hspace{0.25\linewidth}
    \begin{minipage}[c]{0.25\linewidth}
         Ours
    \end{minipage}
    \caption{\textbf{Diffusion illusions.} We generate overlaid optic illusions with SDS and our method. While SDS suffers from color artifacts, our methods produce more details and proper color.}
    \label{fig:illusion}
\end{figure}

\paragraph{Illusion Generation.} Prior works have shown that diffusion models can be leveraged to generate optical illusions~\cite{geng2023visualanagrams,burgert2023diffusion_illusions}. In these settings, the same image looks semantically different when transformed. To use the diffusion model sampling process, a previous study shows that the transformation has to be orthogonal~\cite{geng2023visualanagrams}. However, there remain interesting illusions that are not formed by orthogonal transformation. One such is the rotation overlays. Given a base and a rotator image, by composing the base image with the rotator image at different angles, rotation overlays use two images to display four images. 
As such composition is not defined by an orthogonal matrix, the existing method~\cite{burgert2023diffusion_illusions} employs SDS to optimize the base and rotator images. 
Such a method suffers from the over-saturation problem, as shown in Figure~\ref{fig:illusion}. 
We show that our method can generate such optical illusions with better visual quality.




\begin{figure}[t!]
    \begin{subfigure}{0.62\textwidth}
        \centering
        \includegraphics[width=\textwidth]{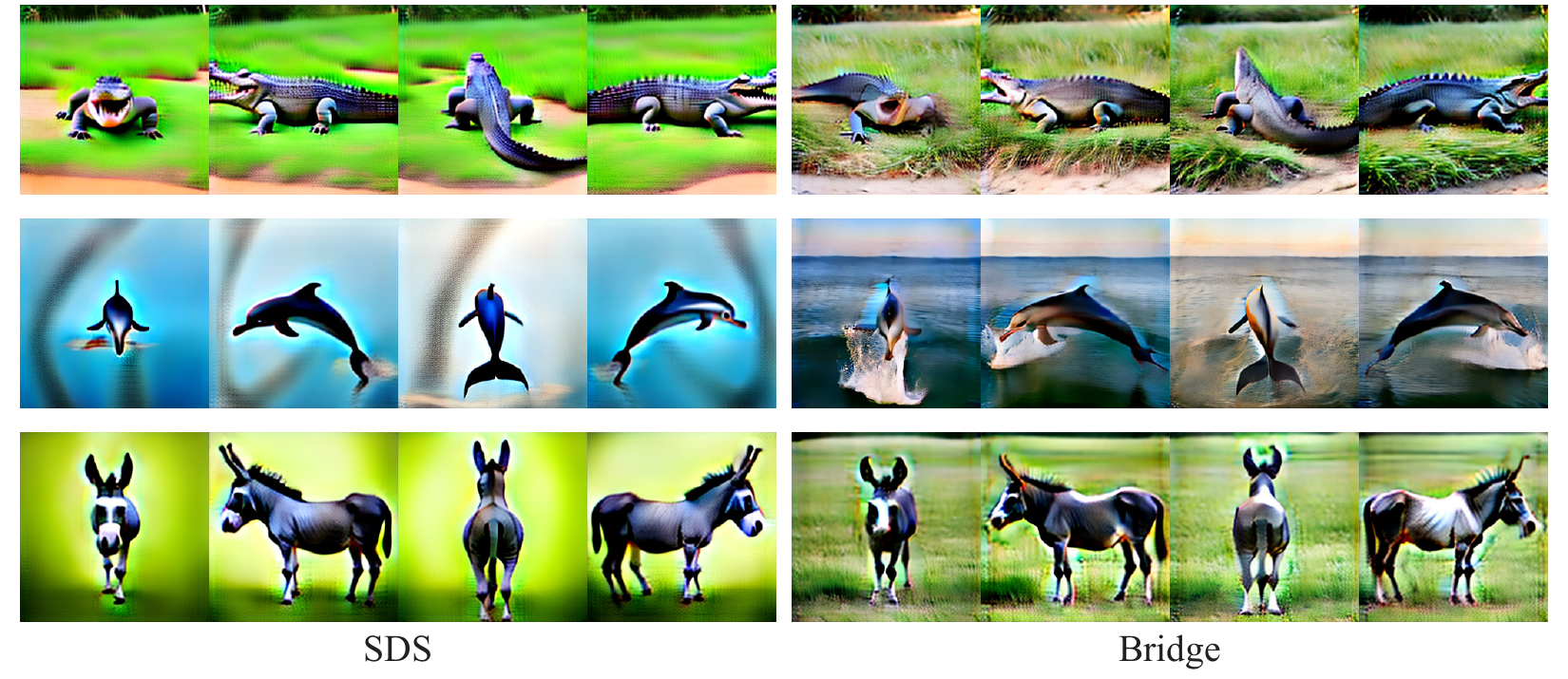}
        \caption{MVDream~\cite{shi2024mvdream}}
    \end{subfigure}%
    \hfill
    \begin{subfigure}{0.38\textwidth}
        \centering
        \includegraphics[width=\textwidth]{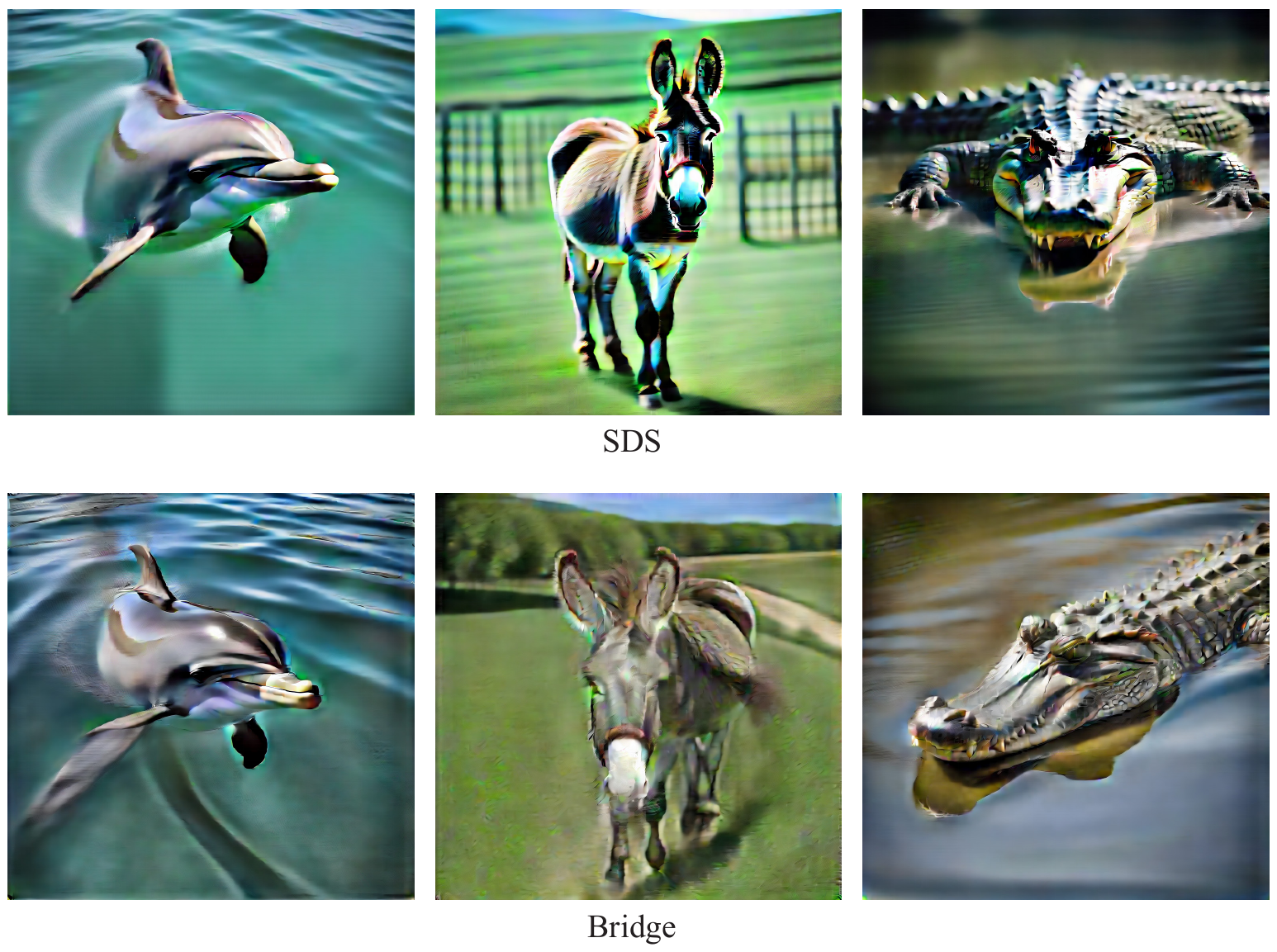}
        \caption{SDXL~\cite{podell2023sdxl}}
    \end{subfigure}
    \caption{\cameraready{\textbf{Comparison of SDS and ours with MVDream~\cite{shi2024mvdream} and SDXL~\cite{podell2023sdxl}.} We compare SDS with our two-stage process in two new settings (MVDream and SDXL). The two-stage process produces more natural colors and realistic details.}}
    \label{fig:mvdreamsdxl}
\end{figure}

\subsection{Additional Qualitative Results}

\cameraready{\paragraph{Additional text-to-image results.} We explore our proposed method across different base models in text-to-image experiments, including MVDream~\cite{shi2024mvdream} and SDXL~\cite{podell2023sdxl}. Since MVDream denoises four camera-conditioned images jointly, we treat the canvas of four images as a single optimization variable for the SDS gradient. In Figure~\ref{fig:mvdreamsdxl}, we compare the SDS baseline to the proposed two-stage optimization, in which we generate more natural colors and detail. This is especially noticeable in the background around the crocodile and donkey.}

\begin{figure*}[t!]
    \centering
    \vspace{-0.8cm}
    \includegraphics[width=\linewidth]{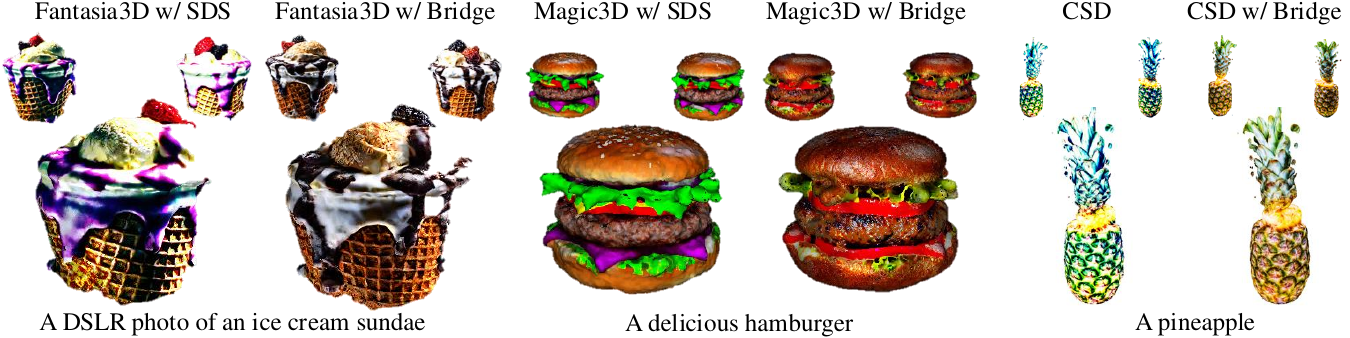}
    \caption{\cameraready{\textbf{Comparison with more text-to-3D baselines.} We apply our two-stage optimization as a drop-in replacement of SDS in Fantasia3D~\cite{Chen_2023_ICCV}, Magic3D~\cite{Lin_2023_CVPR} and CSD~\cite{yu2023textto3d} for texture refinement. We notice that this change greatly improves details and visual quality and reduces SDS artifacts.}}
    \label{fig:more3d}
\end{figure*}

\begin{figure}[t!]
    \begin{subfigure}[b]{\linewidth}
        \includegraphics[width=\linewidth,trim=0 0 0 0,clip]{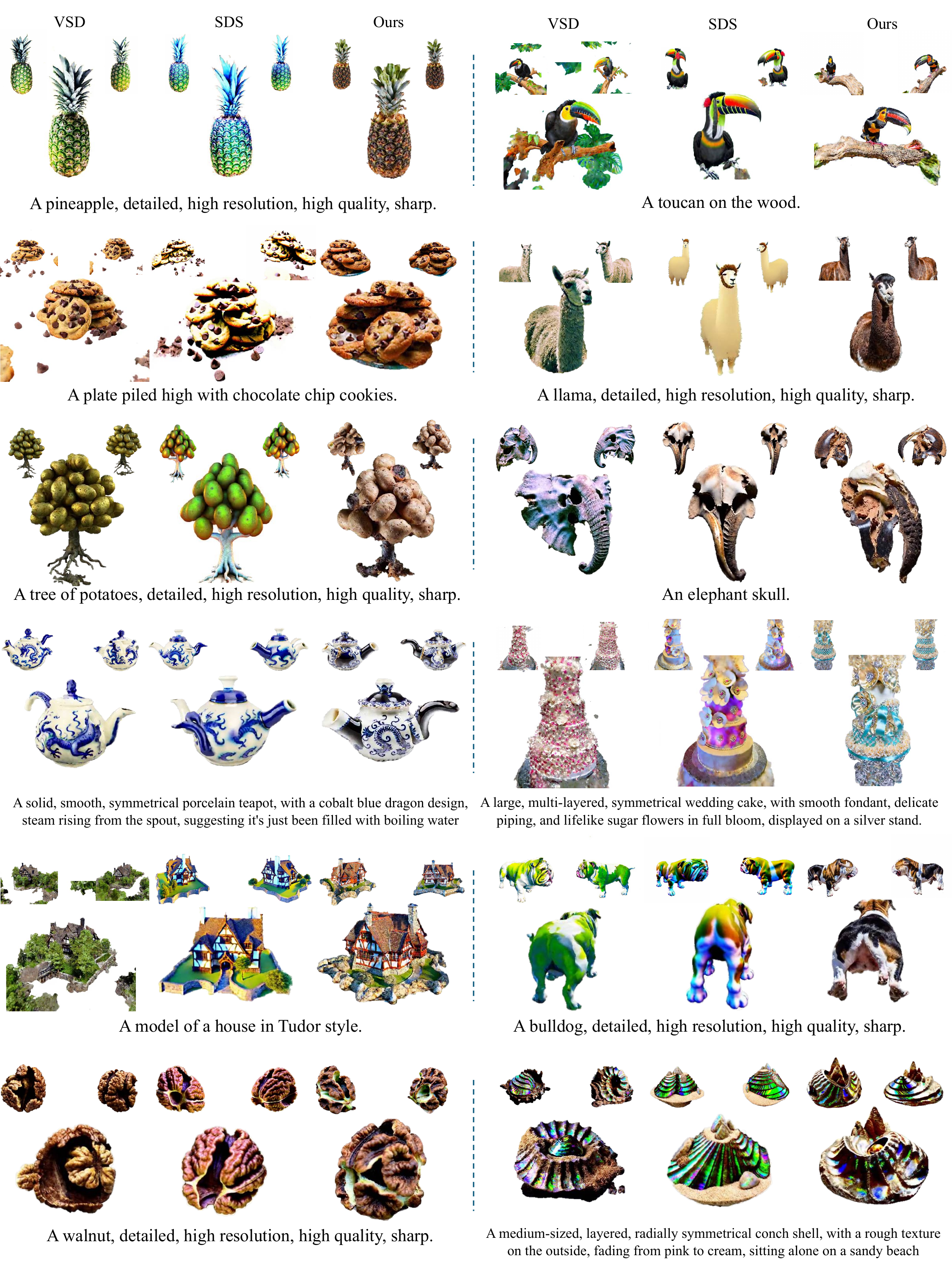}
    \end{subfigure}
    \caption{\textbf{Additional comparison of text-guided NeRF optimization.} We show more examples to compare with different distillation methods, SDS and VSD. }
    \vspace{-2em}
    \label{fig:prolificdreamer_supp4}
\end{figure}

\paragraph{Additional text-guided NeRF optimization results.} For text-guided NeRF optimization comparison against baselines, we show more results in Fig.~\ref{fig:prolificdreamer_supp4}.  
We test on the prompts used in the original paper~\cite{wang2023prolificdreamer} and additional prompts~\cite{wu2023gpteval3d} that we find to be challenging. We notice that SDS often suffers from over-saturation problems. Our method does not require training a LoRA while it can still improve SDS by getting rid of the color artifacts and generating more details.

\cameraready{We also perform comparisons with more competitive baselines. We test with Fantasia3D~\cite{Chen_2023_ICCV}, Magic3D~\cite{Lin_2023_CVPR}, and CSD~\cite{yu2023textto3d} through a drop-in replacement of SDS with our method. Specifically, all three methods optimize a textured DMTet, which is initialized from an SDS-optimized NeRF, using SDS or CSD for 5k or 10k iterations. We replace the SDS or CSD stage of these approaches with the two-stage optimization motivated by our framework. Just like our text-to-3D NeRF experiment, we perform the first stage for 60\% of iterations and the second stage for 40\% of iterations. Note that we keep all the other hyperparameters the same, which were tuned for the baselines, not our method. This replacement leads to the same optimization time as the original methods. For Fantaisia3D and Magic3D, we use threestudio for fair comparison (Magic3D does not have code available) and the default prompts, which are generally believed to work the best with this reimplementation. For CSD, we use the official implementation. As shown in Figure~\ref{fig:more3d}, our method improves the visual quality of all the methods by reducing the oversaturated artifacts of SDS and improving the details.}

\begin{figure*}[t!]
    \centering
    \vspace{-0.2cm}
    \includegraphics[width=\linewidth,trim=0 0 0 832,clip]{figures/additional_exps/prompt_ablation.pdf}
    \caption{\cameraready{\textbf{Ablation study of negative prompts.} We compare SDS results with those from the two-stage optimization (bridge) using our negative prompts and five sets of negative prompts generated by GPT (GPT 1 is the first set, GPT 2 second, GPT 3 third, etc.). All negative prompts produce similar results and outperform the SDS baseline.}}
    \label{fig:ablate_prompt}
    \vspace{-0.2cm}
\end{figure*}

\subsection{Ablation Study}

\cameraready{\paragraph{Ablation study of the negative prompts.} We explore how the choice of negative prompts in our proposed methods affects the optimization. We prompted GPT-4 through ChatGPT a single time to generate alternative negative prompts using the following:}

\cameraready{\textit{Here's a set of "negative prompts" to append to a text-to-image prompt that describe undesirable image characteristics: ", oversaturated, smooth, pixelated, cartoon, foggy, hazy, blurry, bad structure, noisy, malformed" I want to try a variety of them, please brainstorm many of roughly the same length.}}

\cameraready{We produce five variants through these methods as the alternative negative prompts:
\begin{enumerate}
    \item ", washed out, grainy, distorted, flat, smeared, overexposed, undefined, choppy, glitchy, dull"
    \item ", low contrast, jumbled, faint, abstract, over-sharpened, muddy, cluttered, vague, jagged, poor detail"
    \item ", soft focus, muffled, streaky, patchy, ghosted, murky, unbalanced, skewed, mismatched, overcrowded"
    \item ", overbright, scrambled, bleary, blocky, misshapen, uneven, fragmented, obscured, chaotic, messy"
    \item ", dull tones, compressed, smeary, out of focus, unrefined, lopsided, erratic, irregular, spotty, stark"
\end{enumerate}}

\cameraready{We keep other hyperparameters identical and only ablate the negative prompts with the variations. As shown in Figure~\ref{fig:ablate_prompt}, we do not see obvious differences between our prompts and the variants.}

\begin{figure}[t!]
    \begin{subfigure}[b]{\linewidth}
        \includegraphics[width=\linewidth,trim=0 0 0 0,clip]{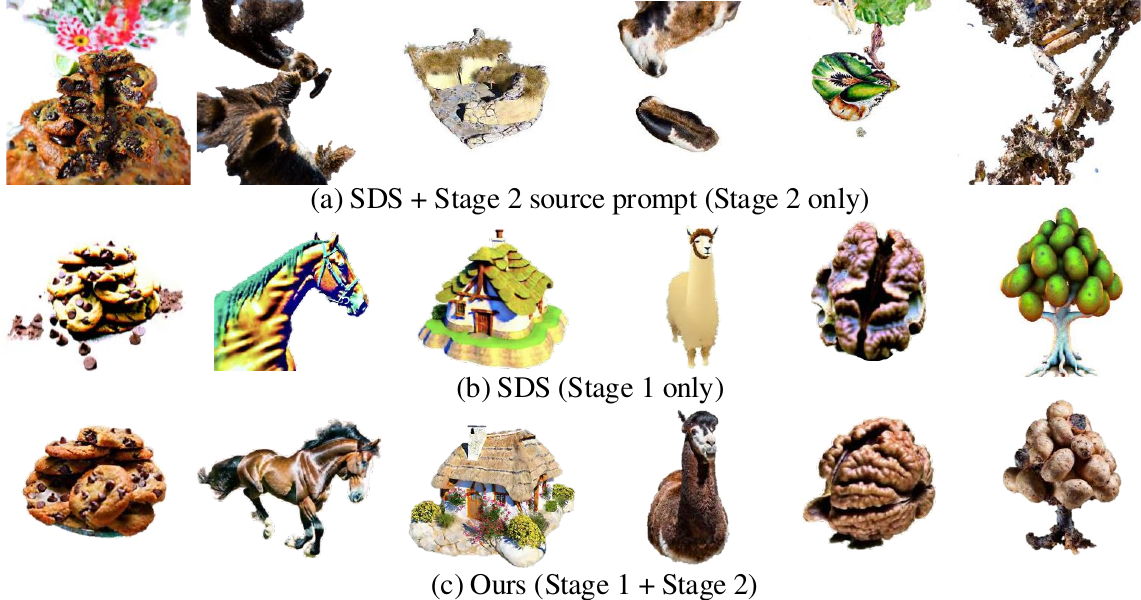}
    \end{subfigure}
    \caption{\cameraready{\textbf{Ablation study of our method without stage 1.} We show directly optimizing with $y_\text{src}$ from the start can undermine the quality of the geometry and produce unnecessary content. }}
    \label{fig:prolificdreamer_ablate}
\end{figure}

\cameraready{\paragraph{Ablation study of stage 2.} Instead of switching to stage 2 during the optimization process, we ablate with starting without any SDS optimization from the beginning. That is, we always use the $y_\text{src}$ with the descriptors ``, \texttt{oversaturated, smooth, pixelated, cartoon, foggy, hazy, blurry, bad structure, noisy, malformed}''. As shown in Figure~\ref{fig:prolificdreamer_ablate}, this makes it hard to generate the proper geometry even though the local texture looks reasonable and is inclined to produce excessive details that are not described by the texts. We suspect that this is because using $y_\text{src}$ increases the mismatching error at the beginning of the optimization process when the initialization does not resemble the target prompt at all.}

\section{Potential Social Impact}
\looseness=-1 We analyze how to use a pre-trained image diffusion as a prior in an optimization setup, necessary for domains such as 3D. On the positive side, these models can empower individuals to make 3D content creation more accessibly without requiring specialized skills. Additionally, professional artists and designers could rapidly prototype and visualize their ideas, accelerating the creative process. On the negative side, the ease of generating visual content could facilitate the spread of misinformation, proliferate biases in the training set and enable the usage of generated content for malicious purposes. In addition, there are ethical concerns regarding the potential for job displacement in industries reliant on traditional art-making skills and the copyright issues appeared in the training dataset.
\clearpage